\documentclass[
]{ceurart}

\usepackage{listings}
\usepackage{lipsum}
\usepackage{graphicx} % Allows including images
\usepackage{tabularx}
\usepackage{booktabs}
\usepackage{amsmath}
\usepackage{amssymb}
\usepackage{booktabs}
\usepackage{array}

\usepackage{pgfplotstable}
\pgfplotsset{compat=1.16}
\usepackage{amsmath}
\usepackage{amssymb}
\usepackage{array}
\usepackage{booktabs}
\usepackage{float}

\usepackage{caption}
\usepackage{algorithm}
\usepackage{algpseudocode}
\usepackage{amsmath}
\usepackage{amssymb}
\usepackage{array}
\usepackage{booktabs}
\usepackage{tikz}
\usepackage{pgfplots}
\usepackage{filecontents}
\begin{document}

\copyrightyear{2024}
\copyrightclause{Copyright for this paper by its authors.
  Use permitted under Creative Commons License Attribution 4.0
  International (CC BY 4.0).}

\conference{LNSAI 2024: First International Workshop on Logical Foundations of Neuro-Symbolic AI, August 05, 2024, Jeju, South Korea}

\title {Abductive Symbolic Solver on Abstraction and Reasoning Corpus}

\tnotemark[1]

\author[1]{Mintaek Lim}[%
email=victorlim@gist.ac.kr,
]
\fnmark[1]

\author[1]{Seokki Lee}[%
orcid=0009-0001-4070-6927,
email=sklee1103@gm.gist.ac.kr,
]

\fnmark[1]

\author[1]{Liyew Woletemaryam Abitew}[%
email=woleteml@gm.gist.ac.kr,
]
\fnmark[1]

\author[1]{Sundong Kim}[%
orcid=0000-0001-9687-2409,
email=sundong@gist.ac.kr,
]
\cormark[1]

\address[1]{Gwangju Institute of Science and Technology}

\fntext[1]{These authors contributed equally.}

\cortext[1]{Correspondance to Sundong Kim (sundong@gist.ac.kr)}

\begin{abstract}
This paper addresses the challenge of enhancing artificial intelligence reasoning capabilities, focusing on logicality within the Abstraction and Reasoning Corpus (ARC). Humans solve such visual reasoning tasks based on their observations and hypotheses, and they can explain their solutions with a proper reason. However, many previous approaches focused only on the grid transition and it is not enough for AI to provide reasonable and human-like solutions. By considering the human process of solving visual reasoning tasks, we have concluded that the thinking process is likely the abductive reasoning process. Thus, we propose a novel framework that symbolically represents the observed data into a knowledge graph and extracts core knowledge that can be used for solution generation. This information limits the solution search space and helps provide a reasonable mid-process. Our approach holds promise for improving AI performance on ARC tasks by effectively narrowing the solution space and providing logical solutions grounded in core knowledge extraction.
\end{abstract}


\maketitle

% 1. %%%%%%%%%%%%%%%%%%%%%%%%%%%%%%%%%%%%%%%%%%%%%%%%%%%%%%%
\section{Introduction}
Artificial intelligence nowadays exhibits impressive problem-solving skills in many domains. Though they provide valuable assistance, not all responses make sense due to the hallucination issue and lack of logical stability. According to Pan Lu et al., especially within the category of mathematical reasoning, logical reasoning, and numeric commonsense, AI agents underperformed compared to other areas such as scientific, statistical, and algebraic reasoning. Moreover, the "puzzle test" and "abstract scene" tasks showed averagely the biggest performance gap between current AI models and humans~\cite{lu2023mathvista}. To enhance such weaknesses, various experiments have been conducted on logic and puzzle test datasets~\cite{chollet2019measure,raven2003raven,antol2015vqa, ghosal2024language}. Datasets corresponding to such categories that require complex logical capabilities with visual images are called Visual Reasoning tasks. 

\begin{figure}
  \centering
  \includegraphics[width=0.75\linewidth]{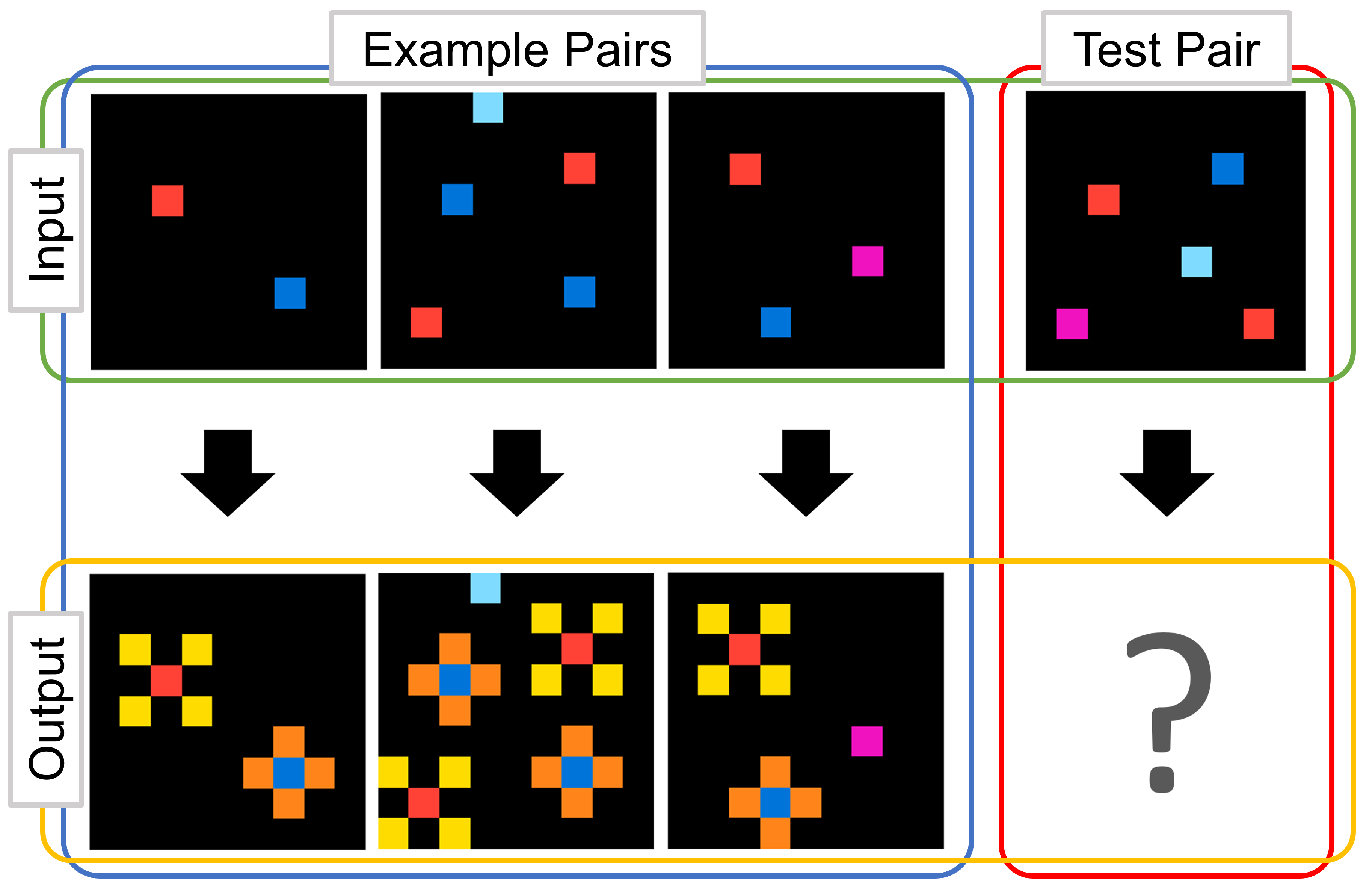}
  \caption{Example ARC task. Solvers are supposed to formulate a pattern that applies to all the given example pairs and then construct an answer with the given test input grid.}
  \label{fig:task}
\end{figure}

As an IQ test is one of the representative measurements of human intelligence, Abstraction and Reasoning Corpus (ARC) was invented by Fran\c cois Chollet to measure the intelligence of an AI~\cite{chollet2019measure}. The ARC dataset has 400 tasks in each training and evaluation set and each consists of multiple numbers of example pairs and a test pair as shown in Figure~\ref{fig:task}. The task is to formulate a pattern that applies to all the example pairs and then construct an answer with the given test input grid. All tasks are created based on four core knowledge priors, which are 1) objectness, including object cohesion, persistence, and its influence via contact, 2) goal-directedness, 3) numbers and counting, and 4) basic geometry and topology~\cite{chollet2019measure}. Due to these characteristics, solutions that have defined domain-specific languages (DSL) have emerged. Unlike other AI techniques, two representative solutions have utilized DSLs to make the essence of each not dissolved into a vector but preserved symbolically. Moreover, the performances have resulted in 1st place in the Kaggle ARC solving competition and ARCathon 2022~\cite{icecuber,hodeldsl}.  Therefore, this research focuses on the symbolic representation of the ARC by applying DSLs and synthesizing DSLs for the solution. 

Since the transformer-based models are considered the best-performing AI, various researchers have challenged solving ARC tasks with texts by providing additional descriptions~\cite{xu2023llms}, applying different prompting skills~\cite{lee2024reasoning}, or estimating hypotheses between the input and output grids~\cite{wang2023hypothesis}. However, such solutions can be improved in the following two ways, 1) by using a symbolic network to generate solutions that are understandable from the human perspective, and 2) by following human thinking processes to make solutions more reasonable and human-like. As humans explain their thoughts to verify their understanding, it is necessary to check both the solution and the answer for reasoning tasks. Thus, this research proposes a symbolic solver that returns understandable and reasonable solutions.

In visual reasoning, humans establish hypotheses based on their observation~\cite{liang2022visual}. Inductive reasoning is well-known as a method to generate general solutions with sufficient observations, however, finding the best solution under limited observations is appropriate with abductive reasoning. Due to such property, the human thinking process of solving the ARC is more likely abductive reasoning. In each pair of the ARC task, the transition between two grids could be represented with multiple hypotheses including 1) what has changed, 2) how or how much it has changed, and 3) why it has changed in such a way. Considering the reason for the transition is the key to this research. In Figure~\ref{fig:task}, four orange pixels appeared around the blue pixel. With only the first pair, it is hard to guarantee a pattern for this task. Observing the second and third pairs provides more clues for formulating a solution. After checking all pairs, the reason for the orange pixel pattern can now be understood, ensuring that the target is blue. In other words, the color is the reason for the pattern not the other fundamental properties like position or counts.

Many previous approaches missed such information and struggled to select a target object to apply the pattern in the solution generation step. By emphasizing the weight of repeated features, we propose an experiment that extracts core knowledge which are the candidate arguments for the solution, and finds common transformations that utilize the extracted information to estimate the result. Our paper's contribution is two-fold, 1) it delineates the conversion of ARC tasks into knowledge graphs and the subsequent extraction of core knowledge from these graphs, and 2) it presents an abductive symbolic solver that utilizes the extracted core knowledge.

% 2. %%%%%%%%%%%%%%%%%%%%%%%%%%%%%%%%%%%%%%%%%%%%%%%%%%%%%%%
\section{Related Works}
\paragraph{Domain Specific Languages (DSL)}
In tackling the ARC challenge, some researchers have designed DSLs by referencing specific ARC tasks and refining them after solving training tasks. While these DSLs prove their systematic stability through successful example pair augmentation based on handcrafted solutions, their adaptability to unseen tasks is limited~\cite{hodeldsl, hodel2024addressing}. Recent studies have explored integrating neurodiversity-inspired methods with computational intelligence through DSLs. One such system, the Visual Imagery Reasoning Language (VIMRL), simulates human mental imagery processes in neurodivergent individuals but struggles to generalize across diverse ARC tasks~\cite{ainooson2023neurodiversity}. Another study uses DreamCoder synthesis to create symbolic abstractions from solved tasks and design a reasoning algorithm, however, this approach heavily relies on previously solved tasks, making it less effective in novel situations~\cite{alford2022neural}.

\paragraph{Graph in ARC}
The paper "Abstract Reasoning with Graph Abstractions (ARGA)" proposed using a graph-based representation to abstract input images into nodes and edges~\cite{Xu2023Graphs}. This method captures spatial and relational information but struggles with the complexity of real-world visual reasoning tasks due to its reliance on predefined graph structures and constraints, limiting its flexibility in diverse scenarios. Lastly, the paper "Mimicking Human Solutions with Object-Centric Decision Transformer" proposed an object-construction algorithm by transforming the ARC grid into a graph to cluster the nodes based on their distances~\cite{park2023unraveling}. Since the aim of the paper is limited to defining an object within only one layer of the graph, the current research gained motivation to expand the graph space by detecting multiple features.

\paragraph{Abductive Reasoning}
Abductive reasoning is a type of logical inference aimed at determining the simplest and most probable explanation from observations. It is used in fields like logistics, design synthesis, and visual reasoning~\cite{liang2022visual, kovacs2005abductive, lu2012abductive, thagard1997abductive}. Liang et al. introduced a task to evaluate machine intelligence in visual scenarios through abductive reasoning. This approach, reflecting human cognition via Observation \(O\) and Explanation \(H\), influenced our understanding of the ARC task~\cite{liang2022visual}. 

\paragraph{Program Synthesis} Program synthesis has shown significant advancements in recent years, particularly in the context of the Abstraction and Reasoning Corpus (ARC). Various approaches have been proposed to tackle the complexities of synthesizing programs that can generalize well from a few examples. For instance, \cite{gulwani2011automating} developed techniques for automating string processing in spreadsheets using input-output examples, laying the foundational work for programming by example (PBE) systems. \cite{chen2018towards} extended this by synthesizing more complex programs beyond domain-specific languages. A notable contribution is the Semantic Interpreter by \cite{gandhi2023natural}, which leverages large language models (LLMs) to translate natural language user utterances into executable programs within a domain-specific language (DSL) tailored for Microsoft Office applications. This approach highlights the effectiveness of combining natural language processing with program synthesis, particularly in productivity software. Building on the idea of leveraging structured domain knowledge, \cite{witt2023divide} introduced a Divide-Align-Conquer strategy for program synthesis. Their method addresses the exponential growth of the search space in program synthesis by decomposing tasks into smaller, manageable subtasks. This approach utilizes structural alignment to guide the search process, significantly improving the efficiency and accuracy of program synthesis in structured domains such as string transformations and visual reasoning tasks in the ARC. By employing analogical reasoning and the Structure-Mapping Theory (SMT), their agent, BEN, outperforms traditional inductive logic programming (ILP) methods, demonstrating the potential of decomposition-driven synthesis in handling complex program generation.

\newpage
% 3. %%%%%%%%%%%%%%%%%%%%%%%%%%%%%%%%%%%%%%%%%%%%%%%%%%%%%%%
\section{Method}
\begin{figure}[h!]
  \centering
  \includegraphics[width=0.8\linewidth]{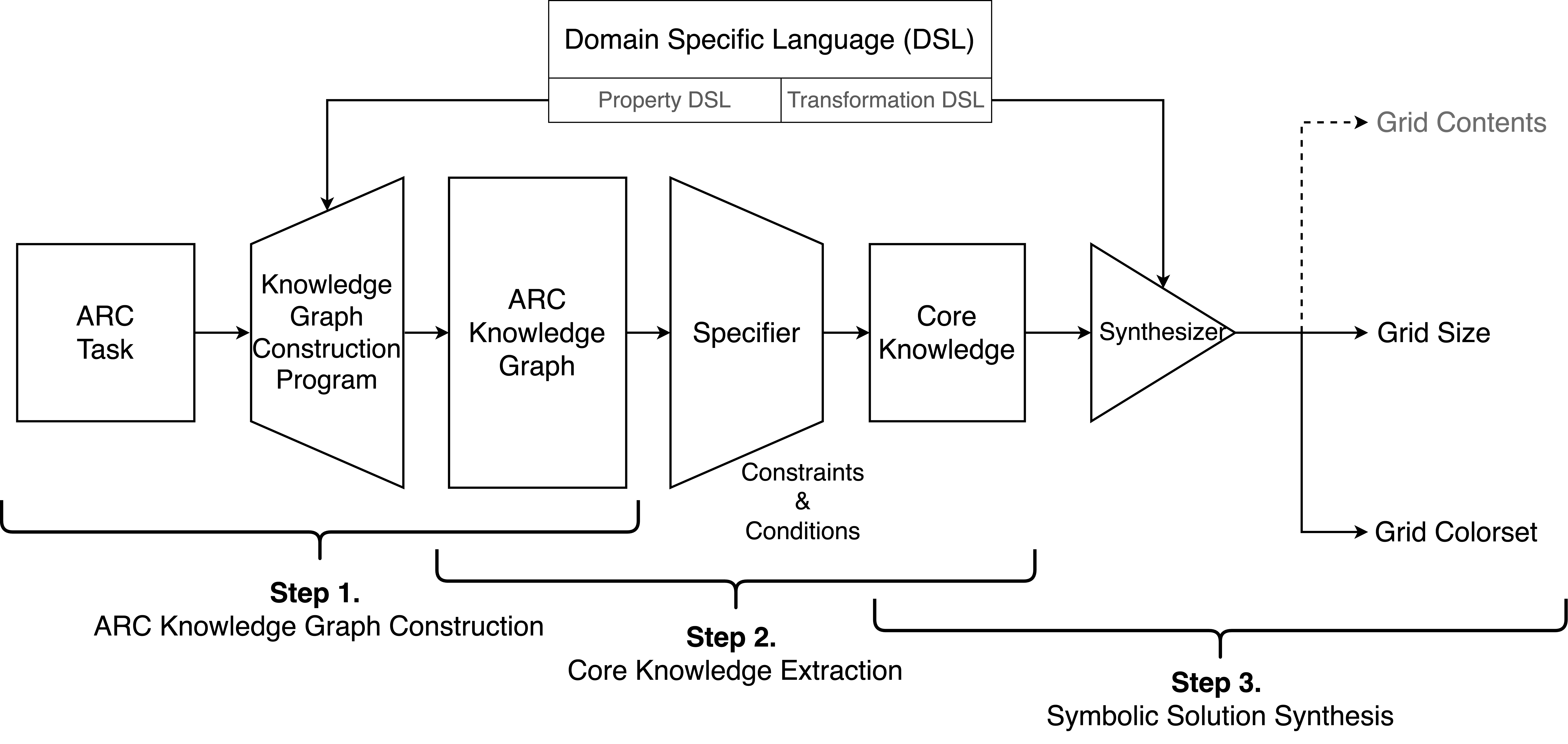}
  \caption{Overall framework of Symbolic ARC Solver. To tackle ARC tasks from the symbolic perspective, the first step involves generating a corresponding knowledge graph using a construction program based on defined Domain Specific Languages (DSL). (Step 1, Chapter~\ref{chap:ARCKG}) Then, extract core knowledge from the knowledge graph using \textit{Specifier}. (Step 2, Chapter 3.2) Since all the ARC tasks consist of multiples of example pairs and a test pair, we define \textit{Specifier} to hold only the repeated conditions that appeared in all example pairs. Lastly, search solutions under given constraints using \textit{Synthesizer}. (Step 3, Chapter 3.3) The information gained from the examples and proposing \textit{Transformation DSL}s limits the solution search space and makes the search feasible.}
  \label{fig:flowchart}
\end{figure}

This chapter will describe the overall structure of solving ARC tasks symbolically with abductive reasoning. The framework shown in Figure~\ref{fig:flowchart} can be divided into three main stages: 1) ARC Knowledge Graph (ARCKG) construction, 2) core knowledge extraction from the knowledge graph, and 3) solution searching using extracted core knowledge. Each of the steps is further described in Sections~\ref{chap:ARCKG}, \ref{chap:Specifier}, and \ref{chap:Synthesizer}, respectively. 

% 3.1. %%%%%%%%%%%%%%%%%%%%%%%%%%%%%%%%%%%%%%%%%%%%%%%%%%%%%%%
\subsection{ARC Knowledge Graph Construction}
\label{chap:ARCKG}
In this knowledge graph construction step, each example pair in the task becomes one unit of ARC Knowledge Graph (ARCKG). For example, a problem in Figure~\ref{fig:task} will have four ARCKGs (three examples and one test pair). ARCKG has four layers in total and it is to organize the nodes and edges well with their origin and characteristics. Based on this 4-layer structure, the construction rule is defined using DSL to apply human understanding to the ARC task and to form a database. DSL in the ARC domain could be categorized into two; \textit{Transformation DSL} and \textit{Property DSL}, and only the \textit{Property DSL}s are used to construct ARCKG. \textit{Transformation DSL}s are used in \textit{Synthesizer} which is explained in Chapter~\ref{chap:Synthesizer}. In the following three sub-chapters, the definition of DSL, the structural frame of ARCKG, and the detailed process of the construction are described.

% 3.1.1. %%%%%%%%%%%%%%%%%%%%%%%%%%%%%%%%%%%%%%%%%%%%%%%%%%%%%%%
\subsubsection{Domain Specific Language Definition}
When humans observe the ARC task, they don't only identify the changes or differences on the surface but also why such changes occurred. According to how Michael Hodel designed his DSL for the ARC~\cite{hodeldsl}, property, and util DSLs are the one that composes the reason for the transformation. In other words, for the complete solution of the ARC, such DSLs are supposed to be preceded before the transformation. Since this research proposes to use a knowledge graph as a source of core knowledge, ARCKG is designed to contain information that could be the key argument of the \textit{Transformation DSL}. Thus, mainly the DSL which represents the property of an object or a pixel is used for the ARCKG construction.

\begin{figure}[tbh!]
    \centering
    \includegraphics[width=1\linewidth]{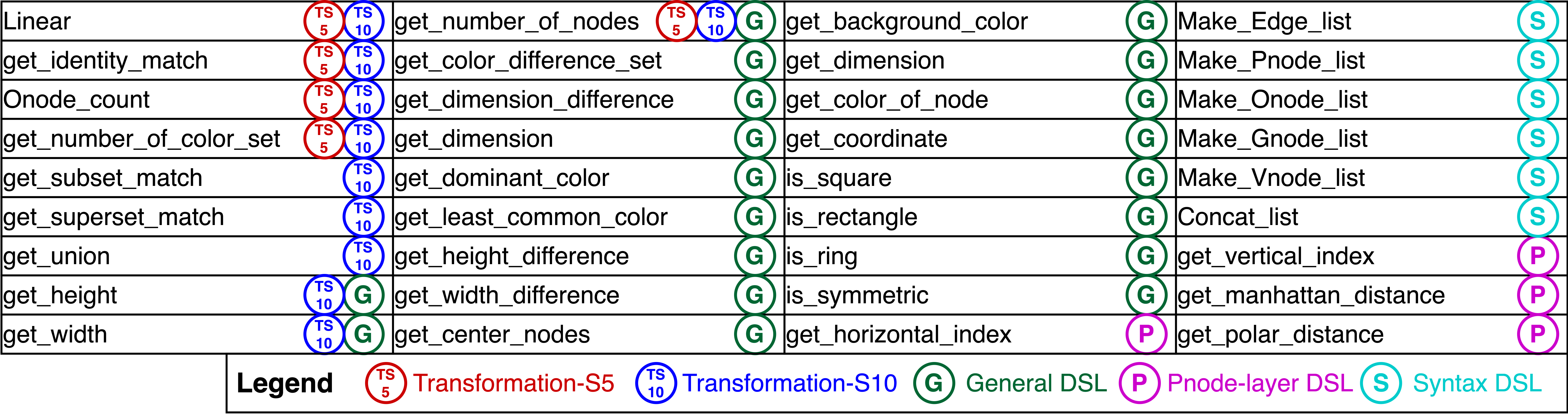}
    \caption{Overview of Domain-Specific Languages (DSLs) and their category tag.}
    \label{Finall_dsl.png}
\end{figure}

\paragraph{DSL Categories - Property DSL} This research proposes DSLs that are classified into two categories based on their purpose. DSLs that symbolize the properties of nodes are referred to as \textit{Property DSL}s and are primarily used to draw edges in the knowledge graph. Refer to the Figure~\ref{Finall_dsl.png} to see what properties are defined. There are several conditions to draw an edge, such as when two nodes have the same property, when a node has a specific property, or when one node is contained within another node by some property. This category is further divided into more specific categories: \textit{General} and \textit{Pnode layer}. The former applies to all layers, generating edges, while the latter applies only to the Pnode layer. \textit{Syntax DSL}s handle the syntactical elements of DSLs and form the backbone of constructing the knowledge graph. They, in turn, are divided into DSLs for generating edges, creating nodes, and combining the two lists, ultimately resulting in the knowledge graph being stored in the form of \textit{nodelist} and \textit{edgelist}.

\paragraph{DSL Categories - Transformation DSL}
\textit{Transformation DSL}s are utilized in the symbolic ARC solver and play a role in predicting the answer by applying transformations to the given nodes. Some of them belong to both \textit{Property DSL} and \textit{Transformation DSL} simultaneously, and the detailed classification is shown in Figure~\ref{Finall_dsl.png}. The reason is due to the ARCKG structure that is defined to have only four types of nodes. The argument of a \textit{Transformation DSL} is

\begin{figure}[tbh!]
    \centering
    \includegraphics[width=0.65\linewidth]{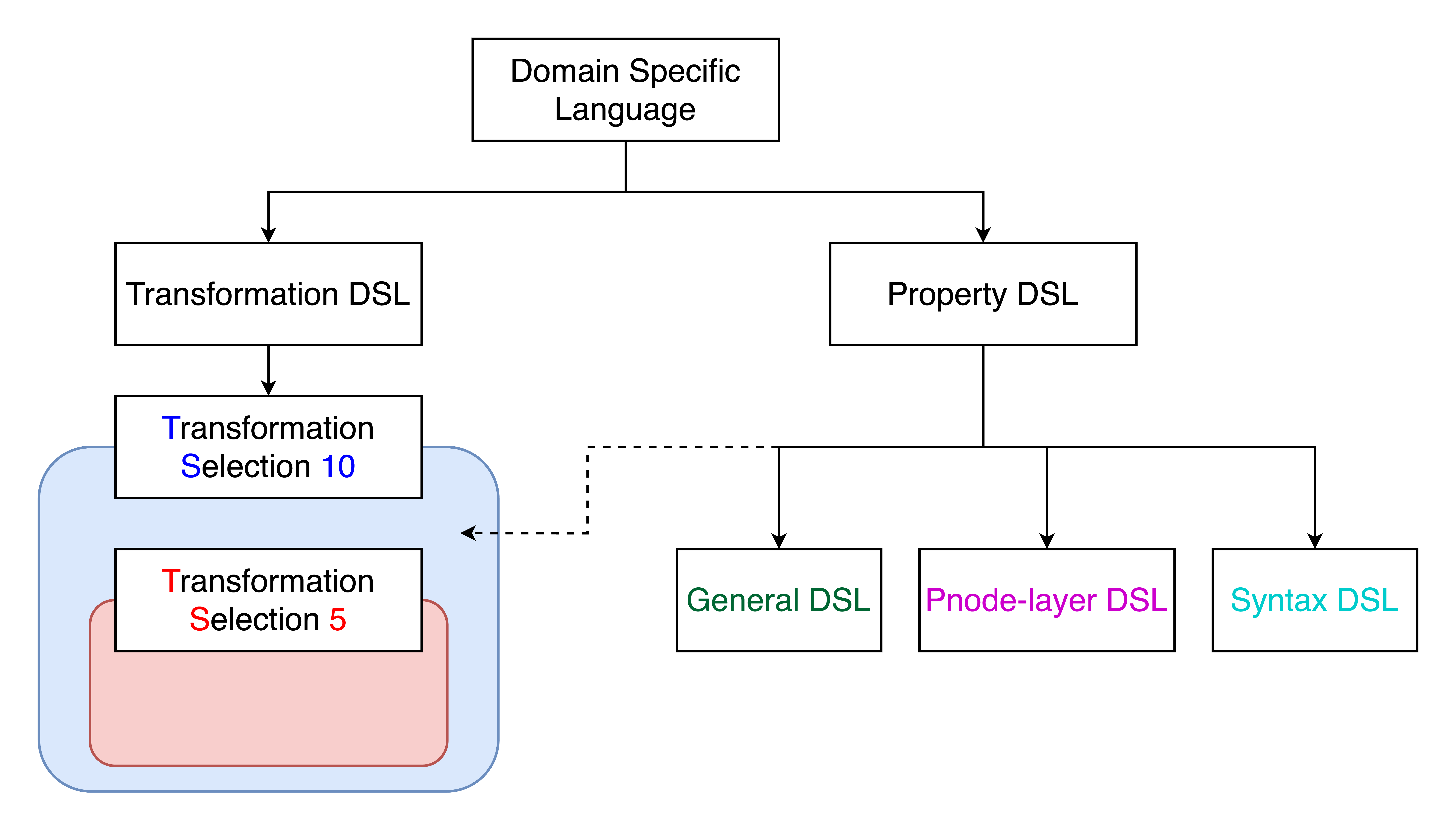}
    \caption{The taxonomy of the Domain-Specific Language (DSL). The terms \textit{Transformation DSL} and \textit{Property DSL} are equivalent to the DSL used in \textit{Synthesizer} and ARCKG construction respectively. In particular, \textit{Transformation DSL}s do not follow the traditional ones, such as move, flip, or rotate due to the experimental setup of this research. \textit{Transformation Selection 10} (\textit{TS10}) contains a selection of suitable DSLs for the experiment and \textit{TS5} is a subset of it. \textit{General DSL} takes the majority of the \textit{Property DSL} and represents the characteristics of the object. Similarly, \textit{Pnode-layer DSL} only appears in Pnode-layer and forms the fundamental feature of object forming. \textit{Syntax DSL} contains node and edge list generation functions to store the information in the form of \textit{NodeList} and \textit{EdgeList}. }
    \label{DSL_definition.png}
\end{figure}

\newpage
\paragraph{Data Types} In the realm of Domain-Specific Language (DSL), data types form the backbone of how information is represented, manipulated, and interpreted. Table~\ref{tab:type} provides an overview of the key data types utilized in our DSL, each tailored to facilitate the unique requirements of nodes and their symbolic relationships.

\begin{table}[ht!]
    \centering
    \caption{Description of Data Types Used in creating DSLs }
    \label{tab:type}
    \begin{tabular}{>{\itshape}l p{10cm}}
        \toprule
        \textbf{Data type} & \textbf{Description} \\
        \midrule
        Pnode & Represent a single pixel in the grid. Stores grid coordinates. \\
        Onode & Represent objects in the grid formed by a collection of Pnode \\
        Gnode & Represent the entire grid holding Pnode and Onode as one node. \\
        Vnode & Represent a pair of input and output into one node that holds two Gnodes. \\
        Xnode & Represent any type of node above. \\
        Edge & Represent relationship between Pnode, Onode, Gnode, and Vnode (provides connection in the graph). \\
        Color & Represent a color of pixel by integer value. \\
        NodeList & Represent a list of nodes. \\
        EdgeList & Represent a list of edges. \\
        Coordinate & Is used to represent coordinates which is a tuple of two integer values. \\
        ColorSet & Is used to hold collections of color. \\
        \bottomrule
    \end{tabular}
\end{table}

% 3.1.2. %%%%%%%%%%%%%%%%%%%%%%%%%%%%%%%%%%%%%%%%%%%%%%%%%%%%%%%
\subsubsection{ARC Knowledge Graph Structure Definition}
The original ARC data is provided in the form of a two-dimensional array, where each element of the array contains information corresponding to colors, ranging from 0 to 9. Therefore, it is challenging for machines to understand and infer rules from this data due to its limited information content. Thus, we propose a method to convert the 2D grid into a knowledge graph that captures information perceived by humans when viewing ARC problems. The knowledge graph is formed as units of one input-output example pair. A single knowledge graph consists of four layers, each characterized by the attributes of the nodes included in it. When representing the original ARC task's example pairs as \(Task = \{ (I_1, O_1), (I_2, O_2), ..., (I_n, O_n) \} \) the corresponding knowledge graphs are expressed as \( ARCKG = \{ g_1, g_2 ..., g_n \} \), where \( g_n \) is further represented as \( g_n = \{ NodeList_n, EdgeList_n \} \). Each \(NodeList_n\) in \( g_n \) is a data structure containing all nodes found in the four layers, and \(EdgeList_n\) is a data structure containing all edges found in the respective knowledge graph. The detailed description of each of the four layers is as follows.

\begin{itemize}
    \item \textbf{Pnode layer:} This first layer converts each pixel into a single node named \(Pnode\) and captures the relationships between these \(Pnodes\), representing them as edges.
    \item \textbf{Onode layer:} This second layer contains nodes representing sets of one or more pixels forming objects. It captures the relationships between objects as edges. Nodes in this layer, which is named \(Onode\), are connected to the \(Pnodes\) with edges.
    \item \textbf{Gnode layer:} This third layer represents the entire input or output grid as a single node named \(Gnode\). Nodes in this layer are connected to all nodes in the lower layers including the first and second with edges.
    \item \textbf{Vnode layer:} This fourth layer combines the input and output grid into a single node. Each example pair is ultimately represented by one fourth-layered \(Vnode\), which is connected to two \(Gnode\)s from the third layer through edges.
\end{itemize}

\begin{figure} [ht!]
  \centering
  \includegraphics[width=0.9\linewidth]{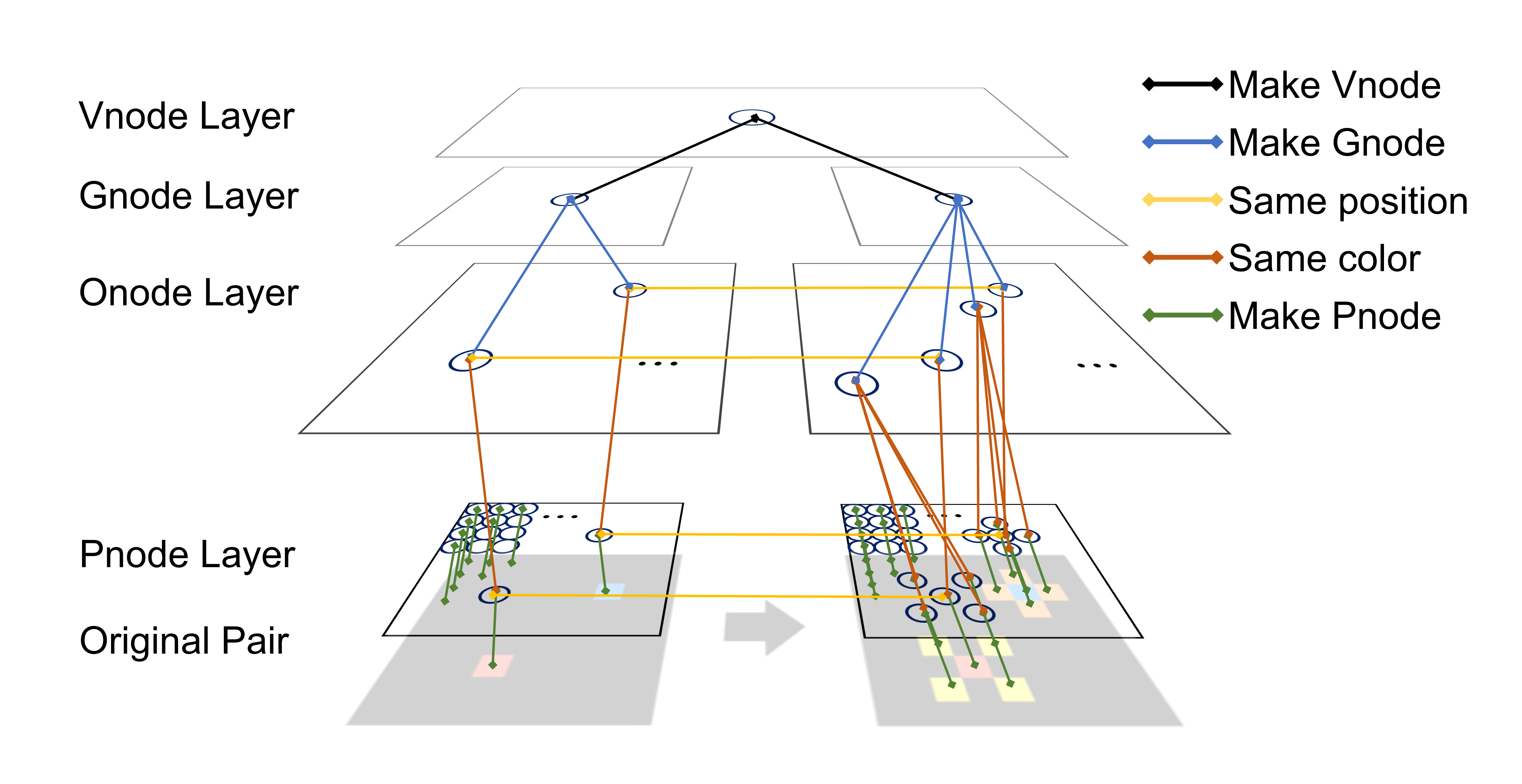}
  \caption{An example of a straightforward, and almost backbone-structured knowledge graph of the first pair of Figure~\ref{fig:task}. In practice, the ARCKG generated by Algorithm~\ref{alg:KgConstruction} can contain up to millions of edges. The graph consists of four layers, with edges freely drawn between layers as well as between input and output by the \textit{Property DSL}. The yellow edges represent connections between two nodes at the same position. The other (black, blue, green) indicate edges signify that nodes in the lower layer constitute nodes in the upper layer.}   
  \label{fig:KG}   
\end{figure}

\newpage
% 3.1.3. %%%%%%%%%%%%%%%%%%%%%%%%%%%%%%%%%%%%%%%%%%%%%%%%%%%%%%%
\subsubsection{ARC Knowledge Graph Construction Program}
Algorithm~\ref{alg:KgConstruction} is an example pseudo-code for building an ARCKG using defined DSLs and graph structure. This program takes a task as input and returns the corresponding knowledge graphs. It consists of two stages: one for generating node lists from the grid and another for creating edges. The algorithm is composed of a nested loop structure. The outer loop iterates over each example pair of the input task. Then, it iterates over the input and output grids of each pair to create node lists. At the end of line 7, two node lists are generated as a result of lines 3 to 7, named \(node\_list\_input\) and \(node\_list\_output\), respectively. Lines 8 to 9 merge these two node lists and create the very top-layer Vnode, appending it to the \(node\_list\). The loop from lines 10 to 12 applies all possible \(Property\_DSL\) to the \(node\_list\), drawing edges using \(Make\_Edge\_list\).

\begin{algorithm}[ht!]
    \caption{ARC Knowledge Graph Construction Program}
    \label{alg:KgConstruction}
    \begin{algorithmic}[1]
        \Require ARC Task
        \State $KG \leftarrow$ empty set
        \For{each $pair$ in $Task$}
            \For{each $grid$ in $pair$}
                \State $node\_list \leftarrow Make\_Pnode\_list(grid)$
                \State $node\_list \leftarrow Make\_Onode\_list(node\_list)$
                \State $node\_list \leftarrow Make\_Gnode\_list(node\_list)$
            \EndFor
            \State $node\_list \leftarrow Concat\_list(node\_list\_input, node\_list\_output)$
            \State $node\_list \leftarrow Make\_Vnode\_list(node\_list)$
            \For{each $Property\_DSL$}
                \State $edge\_list \leftarrow Make\_Edge\_list(node\_list, Property\_DSL)$
            \EndFor
            \State add $(node\_list\_pair, edge\_list)$ to $KG$
        \EndFor
        \State \Return $KG$
    \end{algorithmic}
\end{algorithm}

\newpage
% 3.2. %%%%%%%%%%%%%%%%%%%%%%%%%%%%%%%%%%%%%%%%%%%%%%%%%%%%%%%
\subsection{Core Knowledge Extraction}
\label{chap:Specifier}
In this step, the goal is to extract information that is considered useful for the solution. A unit named \textit{Specifier} takes a knowledge graph as input and returns objects that satisfy the constraints. It plays a role in filtering out relatively less helpful knowledge graph components to narrow down the search space in \textit{Synthesizer}. The conditions of this filter are gathered by analyzing all the given example pairs. Since the ARC is a few-shot task, humans are supposed to formulate a solution from given pairs and apply it to the test grid after only observing a single grid. Conversely, the solution must be appropriately applied to all the examples. Therefore, the solution that we are aiming to find is the intersection of possible solutions of all the given pairs. Moreover, the components of the solution could be found from the ARCKGs established in the previous step, by counting the nodes with identical properties. The following sub-chapters describe the concept of the \textit{Specifier} unit and how it operates differently on example pairs and a test grid.

% 3.2.1. %%%%%%%%%%%%%%%%%%%%%%%%%%%%%%%%%%%%%%%%%%%%%%%%%%%%%%%
\subsubsection{Specifier}
\textit{Specifier} is designed to select candidate objects from the test input grid and by doing so, the afterward solution search space decreases substantially. The object selection in the test grid must be done based on a rule, driven from given example pairs. For instance, the task in Figure~\ref{fig:task} has three example pairs and we can detect the grid changes around a specific pixel. Since the red and blue pixels appear in all three pairs, those pixels in the test grid can become a candidate component for the solution. Accordingly, due to the changes in the grid around those pixels also appearing three times, the solution is concluded to apply such transformations with the corresponding target pixels. Here, in the task given with \(n\) example pairs, selecting the objects, features, or changes observed \(n\) times is critical in \textit{Specifier}. Abductive reasoning in one sentence is to make the best prediction from incomplete observations. Suppose there always is an absolute solution in every ARC task. Due to the incompleteness of the ARC said by the creator Fran\c cois Chollet, given example pairs may or may not express the rule of the task precisely. Thus, the solution-, object-, and constraint-finding process based on the abduction is employed in this research.

\paragraph{Core Knowledge}
The term "core knowledge" refers to an output of the \textit{Specifier}. The narrow range of meaning relates only to the candidates of objects that satisfy the conditions, while the wider range of meaning includes their intrinsic properties. On the surface, \textit{Specifier} unit appears to return only the object on the grid. In contrast, since an object is equivalent to a node in ARCKG, edges that either originated from or ended with the node are also the information on the table. By utilizing the features of selected objects, the \textit{specifier} concludes constraints for object selection in the test grid.

% 3.2.2. %%%%%%%%%%%%%%%%%%%%%%%%%%%%%%%%%%%%%%%%%%%%%%%%%%%%%%%
\subsubsection{Train and Test of the Specifier}
\paragraph{Train of the Specifier}
The training phase of the \textit{Specifier} means the constraint update process for specifying the objects during the example pair observation. Specifically, the update begins from the second pair. During the first pair, there are no objects that can be specified for the solution due to the absence of constraint. Thus, the entire objects and the input grid itself which refer to Onodes and a Gnode become the candidates. This set of objects is then exported to the \textit{Synthesizer}. Regardless of whether the \textit{Synthesizer} discovers the solution path, \textit{Specifier} in the following example starts to filter out the object without any feature in common compared to the object candidates from the previous iteration. The conditions of an object are either a property or a relationship with other components which respectively refer to the term feature and edge. From the second iteration, the constraints of the \textit{Specifier} gathered based on the features and edges of the object are modified. Since no ARC task contains less than two example pairs, this abduction process of updating constraints occurs at least once. After the final update in the last iteration, the constraints are fixed and further used for the test phase.

\paragraph{Test of the Specifier}
During the test phase, the module processes the ARCKG of the test grid. Due to the absence of the output grid, some edges that connect nodes across the grid are not considered. The core knowledge should be driven only from half of the knowledge graph. The trained constraints allow \textit{Specifier} to achieve such a goal under the concept of the task. In short, \textit{Specifier} in the test phase searches for nodes that satisfy the conditions gathered from the example pairs and returns candidate components that can be the material for the solution.

% 3.3. %%%%%%%%%%%%%%%%%%%%%%%%%%%%%%%%%%%%%%%%%%%%%%%%%%%%%%%
\subsection{Symbolic Solution Synthesis}
\label{chap:Synthesizer}
In this step, a solution of an ARC task is discovered by synthesizing \textit{Transformation DSL}s and core knowledge driven from the ARCKG by the \textit{Specifier} unit. A module named \textit{Synthesizer} takes the role of searching through all the combination spaces. Since the solution-finding process follows the brute-force search, theoretically it is solvable under the assumption that the provided DSLs completely cover the task. Moreover, as the \textit{Synthesizer} unit exploits the syntheses of \textit{Transformation DSL} from the example pairs when solving the test case, the following paragraph explains the operation based on the train and test phase.

% 3.3.1. %%%%%%%%%%%%%%%%%%%%%%%%%%%%%%%%%%%%%%%%%%%%%%%%%%%%%%%
\subsubsection{Synthesizer}
The \textit{Synthesizer} unit takes core knowledge and \textit{Transformation DSL}s as input and finds the combination of them to be the desired answer. While humans formulate hypothetical solutions and update them during the example pair observation, \textit{Synthesizer} narrows down the number of solutions. Similar to the object node abduction in the \textit{Specifier}, only the solution that is applicable for all the examples remains after the training phase. Further, when the component reaches the test grid, it exploits the exact solution from the train and returns the answer. Figure~\ref{fig:synthesizer} below depicts the initial solution search space of the first example pair of a task.

\begin{figure} [htb!]
  \centering
  \includegraphics[width=1\linewidth]{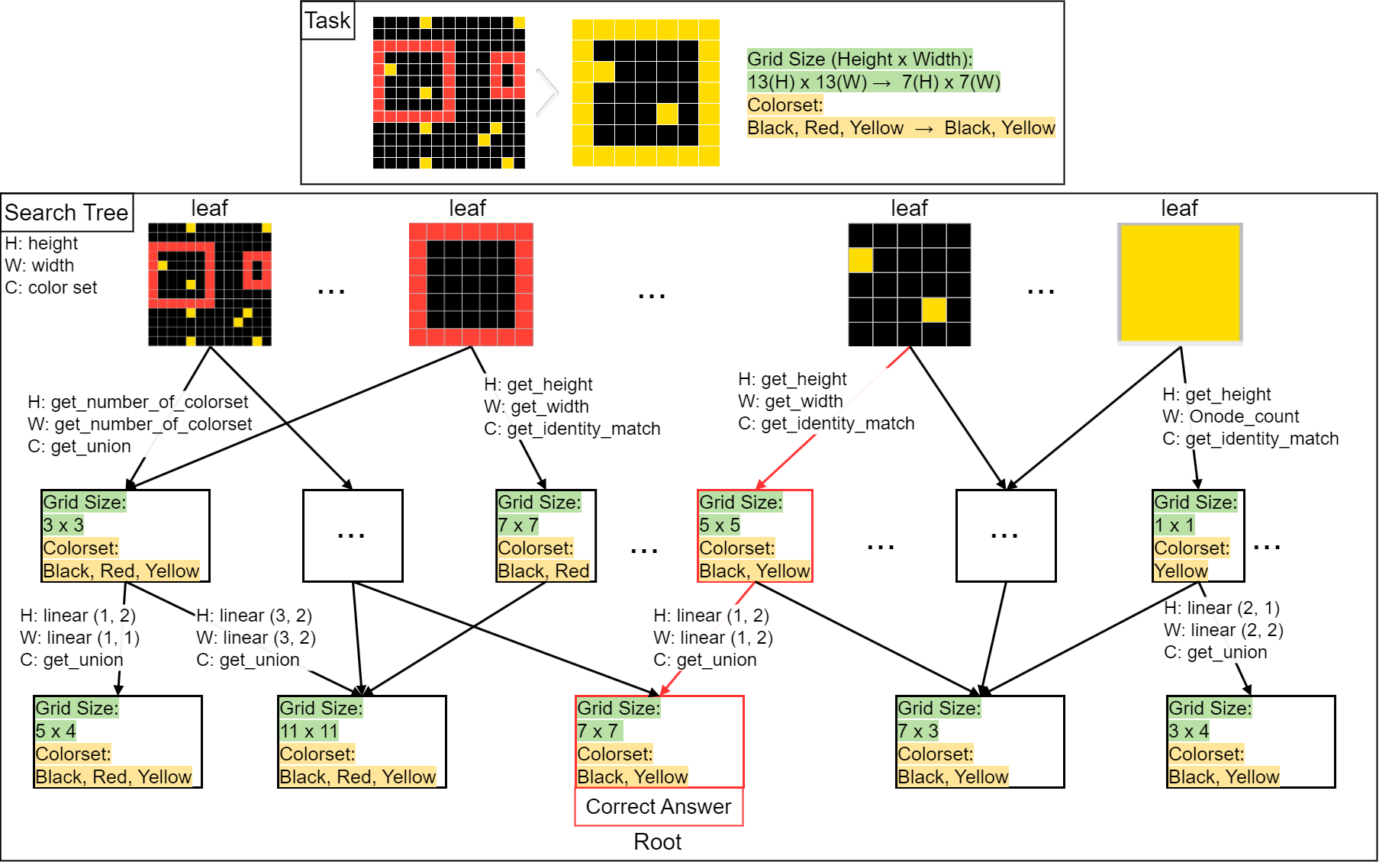}
  \caption{Training session of the \textit{Synthesizer} and its expanded search tree. The task is to find the largest rectangle in the input and change the color to its interior single-pixel color. First, all nodes generated from the input are placed at the top (leaf) of the search tree, with the output node at the bottom (root), commented as Correct Answer in the figure. Then, \textit{Transformation DSL}s are applied to draw paths. This example shows \textit{Synthesizer-10} targeting grid size and color set. Among the DSLs used, \textit{get\_height} returns the height of the node, \textit{get\_width} returns the width, \textit{get\_number\_of\_colorset} returns the number of colors other than the background, and \textit{Onode\_count} returns the number of included objects. The \textit{linear(a, b)} DSL performs the transformation $ax+b$ on the previous value x. \textit{get\_union} returns the union of colors between the previous node and the target node, while \textit{get\_identity\_match} returns the color set of the previous node. The path that reaches the root is highlighted in red, forming a pair with the corresponding leaf.}
  \label{fig:synthesizer}
\end{figure}

% 3.3.2. %%%%%%%%%%%%%%%%%%%%%%%%%%%%%%%%%%%%%%%%%%%%%%%%%%%%%%%
\subsubsection{Abductive Symbolic Solver}
The \textit{Solver}, refers to a union of \textit{Specifier} and \textit{Synthesizer} unit, has an equivalent meaning with the term "abductive symbolic solver" or "symbolic ARC solver" in this paper. It utilizes the concept of abductive reasoning for the learning stage which unfolds in reverse order of inference, starting from the \textit{Synthesizer}. The process begins with each node of the input graph treated as a leaf and extends up to the root of the output grid, exploring all possible paths through the search tree. The edges of the search tree are composed of \textit{Transformation DSL}s, originating from the leaves and branching towards the root by applying each \textit{Transformation DSL}. The search halts when the tree reaches a certain depth, at which point the paths connected to the root become candidates for core knowledge used in the inference stage. At this stage, it identifies all possible \textit{(node, path)} pairs, where the path represents the sequence of \textit{Transformation DSL}s. The term "path" refers to the sequence of \textit{Transformation DSL}s (Domain-Specific Languages) applied within the search tree during the process of abductive reasoning to reach a solution for a given ARC task. Applying this path to the nodes in the pair yields the desired output targeted during the process. An example of the \textit{Synthesizer}'s training can be found in Figure~\ref{fig:synthesizer}, which corresponds to the setup in Section~\ref{sec:setup} and is an example of \textit{Synthesizer-10}. The \textit{Synthesizer} starts with nodes representing parts of the input grid. It applies transformations like \textit{get\_height}, \textit{get\_width}, \textit{get\_number\_of\_colorset}, etc., in sequence. The path through these transformations is formed until the output node, representing the desired solution, is reached.

The \textit{Specifier} generates a function that identifies the minimal features in the knowledge graph that uniquely designate the node, returning them as constraints. 
The objective of this process is to traverse the knowledge graph and find the smallest subset that satisfies the criteria of the given node, such as "same color," "adjacent pixels," and "largest." 
Consequently, the constraint becomes a function that extracts node(s) in the knowledge graph, ultimately generating a hypothesis in the form of a pair \textit{(constraints, path)}. This hypothesis can be applied to all knowledge graphs of the same task by the following method: 
\begin{gather*}
\text{\textit{path(constraints(KG))}} \Rightarrow \text{\textit{prediction}}
\end{gather*}
 It means that by applying the sequence of transformations defined by the "path" to the nodes and information extracted from the knowledge graph (based on the given constraints), the model can generate a prediction or solution for the task. Essentially, the constraints filter and guide the application of transformations, ensuring that only relevant parts of the knowledge graph are used to derive the final prediction.
After obtaining a set of possible hypotheses from the observations of the first example pair, the final solution is adopted through the process of evaluating whether these hypotheses can consistently explain other observations. Due to the nature of ARC problems, observations are highly limited by the number of example pairs and exhibit characteristics of few-shot learning. By applying hypotheses to the given pairs and iteratively selecting only those hypotheses that correctly derive the answers, the remaining hypotheses are adopted as the final solution for this task. This solution ensures that our observations are well explained.

\newpage
\begin{figure}[h!]
  \centering
  \includegraphics[width=0.95\linewidth]{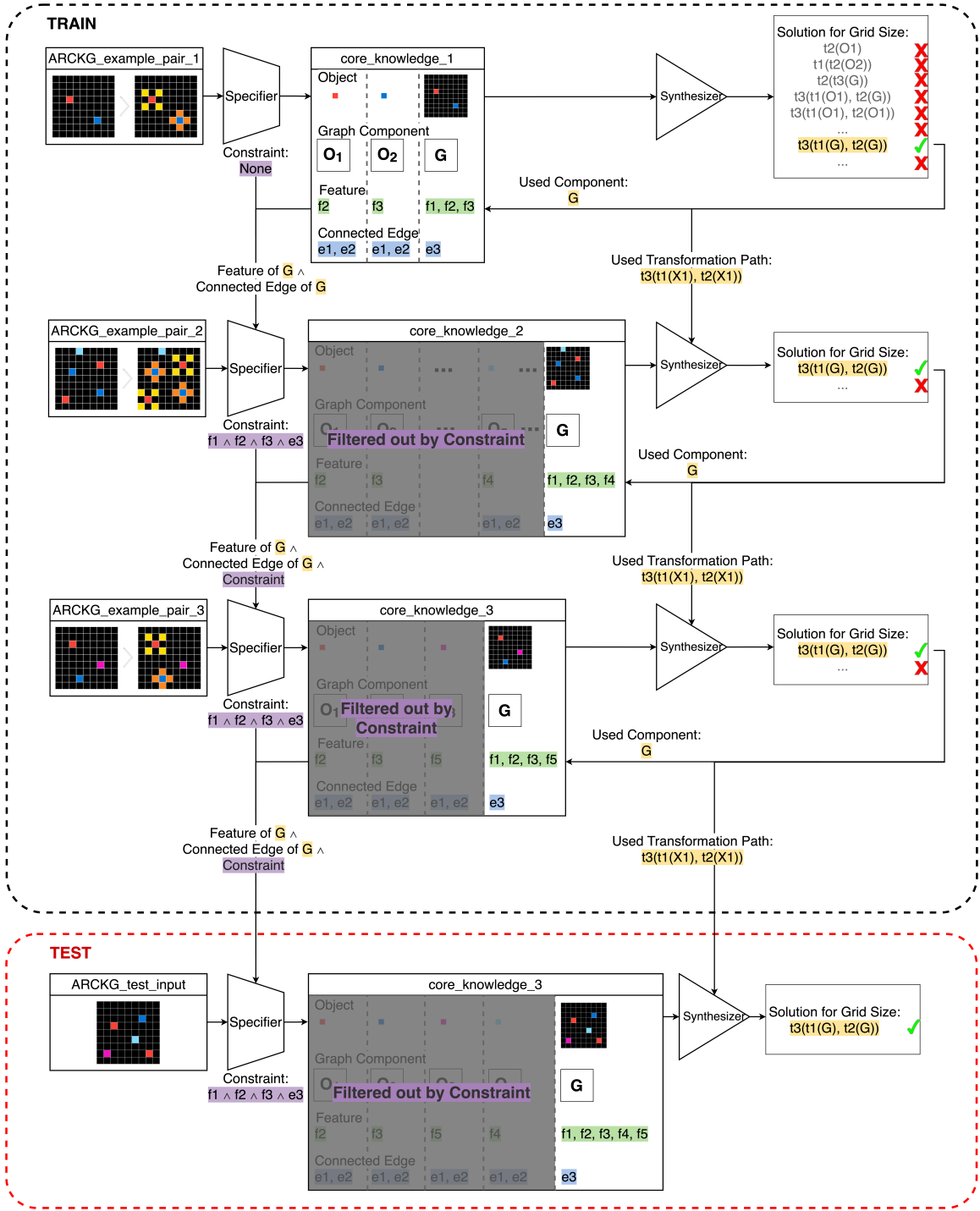}
  \caption{Overall demonstration of proposing symbolic ARC solver. The process consists of two steps, the train phase with given example pairs and the test phase with test input. The starting node indicates the ARCKG constructed using the respective example pair. The initial state of the \textit{Specifier} has no constraint and makes the entire set of detected objects become core knowledge. The \textit{Synthesizer}, can only refer to the graph components which are either Onode or Gnode, thus the candidates do not include other types of node. Throughout the entire combination space, only a few solution paths satisfy the answer and are further utilized for constraint updating. Since the output grid is considered the answer, verification of the result is feasible. The intrinsic core knowledge of the used component, which refers to the feature and connected edge in this diagram, affects the constraint of the following step. The combination of \textit{Transformation DSL}s, the path that leads to the answer, is also transferred to the \textit{Synthesizer} in the next step with generalized form using any node \(X1)\). From the second iteration, \textit{Specifier} and \textit{Synthesizer} follow the conditions of the past. After the training phase, the final conditions are then applied to each unit, and utilize the ARCKG made of the test input grid to yield the answer.}
  \label{fig:example_demo}
\end{figure}

\newpage
% 4. %%%%%%%%%%%%%%%%%%%%%%%%%%%%%%%%%%%%%%%%%%%%%%%%%%%%%%%
\section{Experiment \& Result}
The primary objective of this experiment is to leverage a knowledge graph (KG) and \textit{Domain Specific Languages}s to solve tasks within the Abstraction and Reasoning Corpus (ARC). 
Below are the hypotheses raised in this paper:
\begin{itemize}
    \item \textbf{H1}: The knowledge graphs effectively encapsulate symbolic knowledge, facilitating human-like problem-solving and enhancing performance.
    \item \textbf{H2}: The number of \textit{Transformation DSL}s is positively correlated with the performance of the symbolic ARC solver.

\end{itemize}
\subsection{Experimental Setup}
\label{sec:setup}
To evaluate the performance of the DSL-based symbolic Arc solver, we conducted experiments with two distinct setups: one utilizing a knowledge graph and another without it. This comparison aims to assess the impact of knowledge graphs on the solver's accuracy in predicting the ARC task outputs(grid size and color set).

\paragraph{Target Elements:} The answers (outputs) of ARC problems consist of three elements: 1) the size of the grid, 2) the color set of the grid, and 3) the contents of the grid. Though all three are crucial, predicting and modifying the target values of the first two hold significant importance as they represent steps inherent in human problem-solving of ARC tasks. Therefore, we prioritized these aspects in our experimental setup, focusing primarily on them and enabling the utilization of minimal and straightforward \textit{Transformation DSL}s during the synthesis process. For the color set, all colors appearing in the correct grid must be matched with the predicted value to be considered as the correct answer, while for the grid's size, separate integer values for height and width were predicted.

\paragraph{Approach using Knowledge Graphs:}In this experiment a total of 22 \textit{Property DSL}s were employed to build a graph encapsulating the symbolic information of the grid elements. Based on the transformations defined in the DSLs the solver generates potential solutions followed by the \textit{Synthesizer} selecting the most accurate solutions by leveraging the information from the target node extracted by the \textit{Specifier} from the knowledge graph. 

\begin{figure}[h!]
  \centering
  \includegraphics[width=0.8\linewidth]{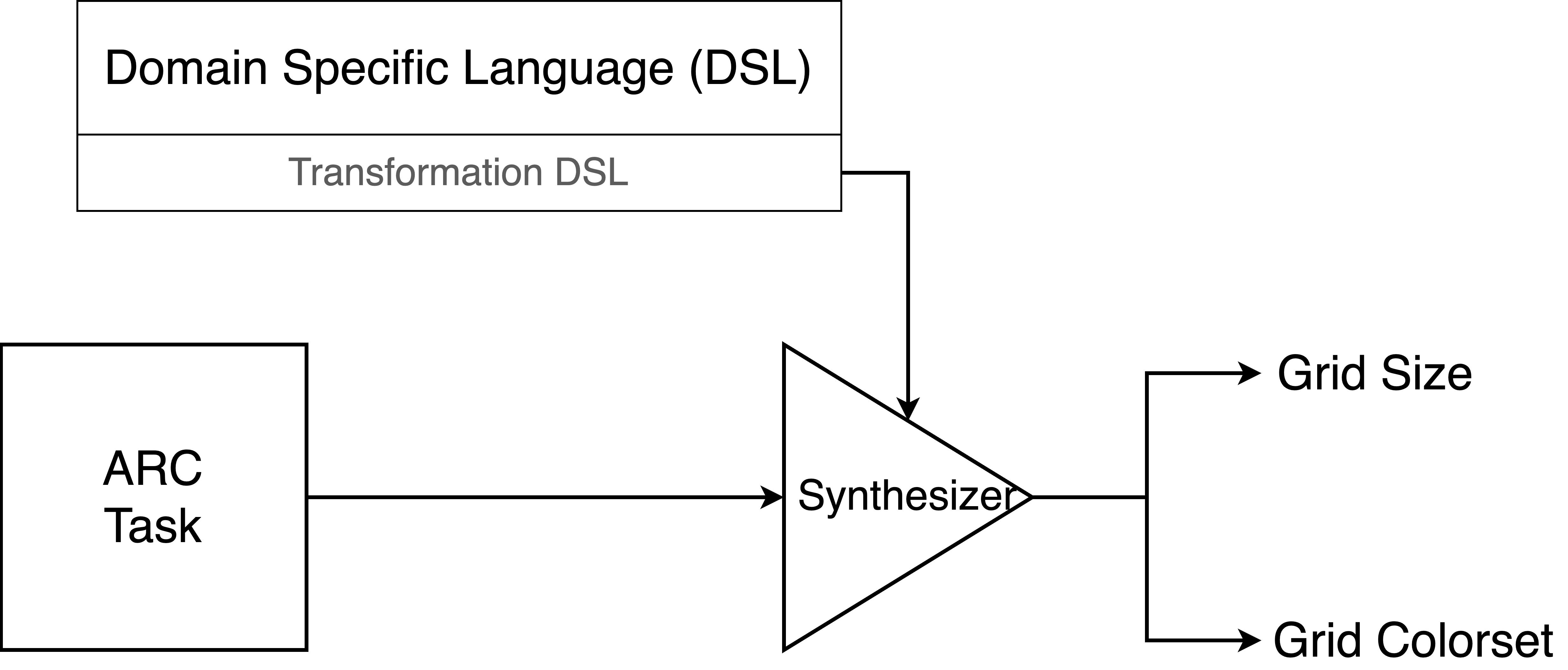}
  \caption{Systematic schema of the experiment without knowledge graph. Since the knowledge graph is not used, the process of graph construction and core knowledge extraction are omitted. Accordingly, only the \textit{Transformation DSL}s are used. }
  \label{fig:woKG}
\end{figure}

\paragraph{Approach without Knowledge Graphs:}This setup is similar to the experiment with the knowledge graph construction, but without the intermediate step of graph construction. The transformations DSLs are directly applied to the grid elements to generate potential solutions. Thus In this experimental setup, no \textit{Specifier} is needed since the goal of a \textit{Specifier} is to extract the unique characters of nodes from the knowledge graph. The overall flow of how we experimented without the knowledge graph is depicted in Figure  ~\ref{fig:woKG}.

\paragraph{Experimental Procedure:}A set of ARC tasks (400 tasks) was selected for the experiments ensuring a diverse range of grid sizes and color sets. For the KG approach graphs were constructed for each task using the 22 \textit{Property DSL}s. Then both solvers (KG-based solvers and non-KG-based solvers) run on the tasks to predict the grid size and color set. The accuracy of the solvers was measured based on the correctness of the grid size(height and width) and color set.
\subsection{Result}

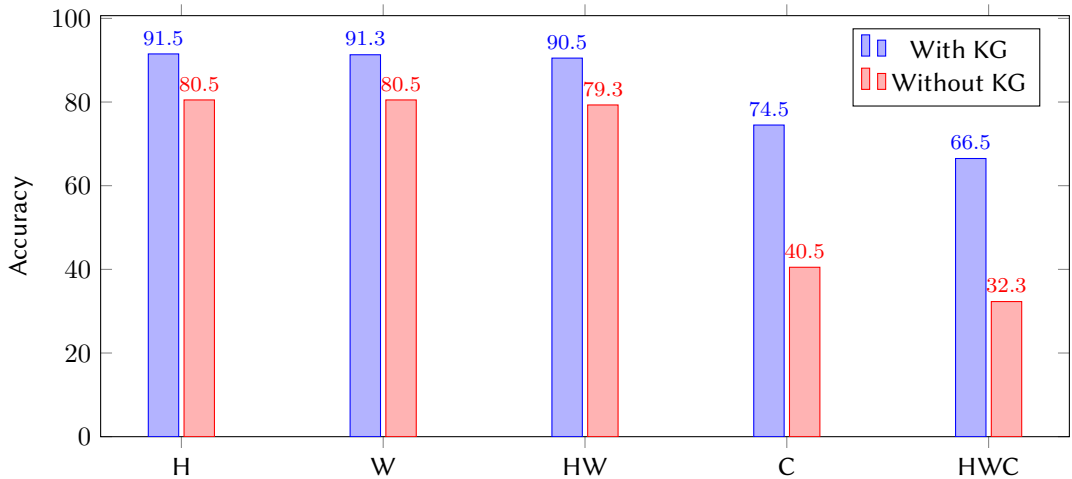
\begin{figure}[h]
    \centering
    \begin{tikzpicture}
        \begin{axis}[
            ybar,
            symbolic x coords={H, W, HW, C, HWC},
            xtick=data,
            ymin=0,
            ylabel={Accuracy},
            legend pos=north east, % Adjusted legend position
            bar width=0.4cm,
            width=0.9\textwidth,
            height=0.45\textwidth,
            nodes near coords,
            nodes near coords align={vertical},
            every node near coord/.append style={font=\scriptsize\color{black}},
        ]
        \addplot table[x=Component,y=With Knowledge graph,col sep=comma] {data.csv}; \addlegendentry{With KG}
        \addplot table[x=Component,y=Without knowledge graph,col sep=comma] {data.csv}; \addlegendentry{Without KG}
        \end{axis}
    \end{tikzpicture}
    \caption{Accuracy score comparison of solver with and without utilizing knowledge graph on each target. Here, "KG" refers to the knowledge graph. The targets assessed are Height (H), Width (W), Color (C), and their combinations: Height and Width (HW), and Height, Width, and Color (HWC).}
    \label{fig:counts}
\end{figure}

\paragraph{Comparison of Solver Performance with and without the Use of Knowledge Graph}
Figure~\ref{fig:counts} presents the accuracy scores of a solver’s performance on different target values, comparing the use of a knowledge graph against not using one. For each target on the x-axis, the solver’s accuracy is consistently higher when utilizing the knowledge graph. In particular, when not utilizing the knowledge graph, a significant decrease in the prediction performance of C and HWC can be observed. This indicates the crucial role of symbolic information contained in the knowledge graph in predicting the color set. The solver achieves nearly perfect accuracy for the H, W, and HW with the knowledge graph. These results confirm that the use of knowledge graphs effectively enhances performance, supporting \textbf{H1} by demonstrating their capability to encapsulate symbolic knowledge and facilitate human-like problem-solving.

\paragraph{Differences in Performance by Size of Synthesizer} 
To explore the relationship between the number of \textit{Transformation DSL}s and accuracy, two \textit{Synthesizer}s of different sizes were prepared, both with a depth limit of 2 for the search tree. The results show that \textit{Synthesizer-10} consistently achieves higher accuracy across all categories compared to \textit{Synthesizer-5}. Notably, in the HWC category, \textit{Synthesizer-10} outperforms \textit{Synthesizer-5} by over three times. These findings support H2, confirming that the number of \textit{Transformation DSL}s is positively correlated with the performance of the symbolic ARC solver. Additionally, this suggests that employing more sophisticated and diverse \textit{Transformation DSL}s enhances the model's accuracy and its potential to predict content.

\begin{table}[ht!]
  \centering
  \caption{The comparison presented here delves into the accuracy scores of solvers utilizing different \textit{Synthesizer} sizes. \textit{Synthesizer-10}, employing 10 \textit{Transformation DSL}s, is contrasted with \textit{Synthesizer-5}, which utilizes only 5. For details on DSL adopted by each \textit{Synthesizer}, see Figure~\ref{Finall_dsl.png}.}
  \label{tab:combined_results}
  \begin{tabular}{@{}lcccccc@{}}
    \toprule
    \textbf{} & \multicolumn{3}{c}{\textbf{Synthesizer-10}} & \multicolumn{3}{c}{\textbf{Synthesizer-5}} \\
    \cmidrule(lr){2-4} \cmidrule(lr){5-7}
    \textbf{} & \textbf{Correct} & \textbf{Incorrect} & \textbf{Accuracy (\%)} & \textbf{Correct} & \textbf{Incorrect} & \textbf{Accuracy (\%)} \\
    \midrule
    \textbf{H} & 366 & 34 & 91.5 & 209 & 191 & 52.25 \\
    \textbf{W} & 365 & 35 & 91.25 & 203 & 197 & 50.75 \\
    \textbf{C} & 299 & 101 & 74.75 & 176 & 224 & 44 \\
    \textbf{HW} & 362 & 38 & \underline{90.5} & 197 & 203 & \underline{49.25} \\
    \textbf{HWC} & 266 & 134 & \underline{66.5} & 84 & 316 & \underline{21} \\
    \bottomrule
  \end{tabular}
\end{table}

\section{Conclusion}
We introduced a framework for ARC problem-solving, integrating knowledge graph conversion and abductive reasoning learning with a symbolic ARC Solver. This approach, inspired by human thought processes, offers systematic, interpretable, and scalable solutions. Leveraging knowledge graphs, we decode ARC tasks symbolically, providing crucial insights for inferring problem rules. Impressively, even with a naive \textit{Synthesizer} using limited \textit{Transformation DSL}s, our framework achieves high accuracy in predicting grid sizes (90.5\%) and color sets (74.5\%). Furthermore, as DSLs increase, we anticipate significant performance improvement, potentially extending to grid content prediction.
\begin{lstlisting}

\end{lstlisting}

\begin{acknowledgments}
  This work was supported by the IITP (RS-2023-00216011, No.~2019-0-01842), AICA (HPC-AI) and the GIST (HPC-AI) funded by the Ministry of Science and ICT, Korea. 
\end{acknowledgments}

\bibliography{sample-ceur}

\end{document}